\documentclass[11pt,a4paper]{article}
\usepackage{naaclhlt2019}
\usepackage{times}
\usepackage{latexsym}
\usepackage{amssymb}
\usepackage{url}
\usepackage{examples}
\usepackage{multirow}
\usepackage{color}
\usepackage{xcolor,colortbl}
\usepackage{graphicx}
\usepackage{amsmath}

\aclfinalcopy

\title{
Extracting Factual Min/Max Age Information from Clinical Trial Studies
}

\author{Yufang Hou$^1$, Debasis Ganguly$^1$, L\'{e}a A. Deleris$^2$, 
Francesca Bonin$^1$\\
$^1$IBM Research, Ireland\\
{\tt \{yhou,debasga1,fbonin\}@ie.ibm.com}\\
$^2$BNP Paribas\\ {\tt lea.deleris@bnpparibas.com}\\
}

\date{}

\begin{document}
\maketitle

\begin{abstract}
Population age information is an essential characteristic of clinical trials.
%However, extracting min/max age values for the most complex intervention group from clinical research papers is a      
%In this paper, we propose a system to extract factual age information (i.e., min age and max age) of the population from clinical research papers.
%In this paper, we focus on extracting min/max age values for the most complex intervention group from clinical research journal articles. 
In this paper, we focus on extracting minimum and maximum (min/max) age values for the study samples from clinical research articles. 
%We view this problem as a reading comprehension task 
Specifically, we investigate the use of a neural network model for question answering to address this information extraction task. 
The min/max age QA model is trained on the massive structured clinical study records from \emph{ClinicalTrials.gov}.
For each article, based on multiple min and max age values extracted from the QA model,  
we predict both actual min/max age values for the study samples and filter out non-factual age expressions.
%We leverage 
%the massive structured clinical study records from \emph{ClinicalTrials.gov} to provide distant supervision for the model. %min/max age value extraction.
Our system improves the results over (i) a passage retrieval based IE system and (ii) a CRF-based system by a large margin when evaluated on an annotated dataset consisting of 50 research papers on smoking cessation.
%We also analyze 

\end{abstract}

\section{Introduction}

Clinical trials are an important source of scientific evidence for guiding the practice of evidence-based medicine.
However, many characteristics of clinical trials are only %exist
reported in the published research articles.
The health service community could benefit from knowledge bases populated with detailed information from clinical trials reported in research articles. With this in mind, clinical information extraction 
aims to extract such information from journal articles that report randomized controlled trials \cite{Kiritchenko2010,byron2016}.

Relevant information about clinical trials can be categorised along: \textit{(i) } trial's population characteristics (e.g. minimum and maximum age of the participants, education level, marital status, health status), \textit{(ii) }intervention methods, both what is being done (e.g. specific drug and dosage, planning sessions, use of an app for daily reporting) and how it is being administered (e.g., where, how often and by whom), and \textit{(iii)} outcome of the study (e.g., \textit{30\% of the population stopped smoking after 6 months}).

%Extracting the key characteristics of different arms from research articles about clinical trials is a challenging task. 
%In this paper, we focus on extracting min/max age values for the most complex intervention group.

In this paper, we focus on extracting population characteristics and in particular minimum and maximum (min/max) age values %of the included study samples from 
associated with the study samples %as described in 
from clinical trials research articles.

Unlike \cite{summerscales}, our aim is to extract information from the full article, rather than only from the abstract, as we have observed that age information is not always described in the abstracts. In our testing dataset consisting of 50 research papers, only nine papers describe the min/max age information in their abstracts.

Naturally, analysing the entire article presents many challenges. 
Our goal is to identify the factual min/max age value information for the persons who actually participated in the clinical trial (see Example \ref{ex:age1} and Example \ref{ex:age2} below). This should be distinguished from non-factual min/max age information (Example \ref{ex:age3} and Example \ref{ex:age4}) and also from min/max age information which is not related to the participants in the study (Example \ref{ex:age5} and Example \ref{ex:age6}).

\begin{examples}
\item \label{ex:age1} Participants were 83 smokers, who were \textbf{18}-\textbf{23} years old and undergraduate students \ldots
%\item \label{ex:age1} Participants were 83 smokers, who were \textbf{18}-\textbf{23} years old and undergraduate students at a university in the Washington, D.C., metropolitan area.
%\item \label{ex:age3} Eligibility for this study included being a student (full or part time), smoking at least 1 cigarette/day in each of the past 7 days, being aged \textbf{18}-\textbf{24} years, and being interested in
%quitting smoking in the next 6 months.
\item \label{ex:age2} participants aged \textbf{18}-\textbf{24} years were randomized to a brief office intervention (n=99) or to an expressive writing plus brief office intervention (n=97).
\item \label{ex:age3} To be included in the study, smokers had to be between the ages of \textbf{18} and \textbf{60} years \ldots
%\item \label{ex:age3} Patients were included if they  between the ages of \textbf{18} and \textbf{60} years \ldots
\item \label{ex:age4} The subjects were eligible for inclusion if they were at least \textbf{18} years of age, reported smoking 10 or more cigarettes per day, \ldots
\item \label{ex:age5} An estimated 23.6\% of young adults aged \textbf{18}-\textbf{24} years are current smokers. 
\item \label{ex:age6} Smoking Dutch youths had in many cases tried their first cigarette at the age of \textbf{11}-\textbf{12} years.
\end{examples}

Our proposed system extracts factual min/max age values of the study samples directly from research articles in PDF format. 
We leverage the massive structured clinical study records from \emph{ClinicalTrials.gov} to provide distant supervision for min/max age value extraction.
Furthermore, inspired by the work on hedging detection on Bioscience domain \cite{light04,kilicoglu08,farkas10}, we explore a list of ``speculation cues'' to filter out non-factual min/max age expressions.
Our system improves the results over \textit{(i)} a passage retrieval based IE system and \textit{(ii)} a CRF-based system by a large margin when evaluated on an annotated dataset consisting of 50 research papers on smoking cessation.

%
%The following of the paper is structured as follows: we first describe related works; then we describe the methodology. In Section~\ref{sec:eval}, we present the experiments and we draw our conclusions in Section~\ref{sec:con}.
%%
%\cite{summerscales} used some heuristic rules to extract min/max age values of the study population from the abstracts. However,
%age information is not always described in the abstracts. In our testing dataset consisting of 50 research papers, %we found that 
%only nine papers describe 
%the min/max age information in their abstracts. Therefore we decide to extract min/max age values from the full articles. 

\section{Related Work}\label{sec:rel}

\subsection{Clinical Information Extraction}
In general, research on information extraction from medical literature is still in its infancy involving a number of limitations, such as lack of common benchmarking datasets, and a lack of general consensus on the class of approaches that are reported to work well on such benchmarks.

Some work has been conducted on supervised approaches for medical information extraction. Multiple studies have concentrated their efforts on medical abstract. In \cite{Kim:2011}, the authors propose a conditional random filed (CRF) classification method for labelling medical abstract sentences according to medical categories, such as outcome, intervention, population. Hansen et al., 2008 \cite{Hansen:2008} developed a Support Vector Machine algorithm for extracting the number of trials participants from medical abstracts, while in \cite{HASSANZADEH2014}, the authors use a machine learning approach for classifying abstract sentences according to the 
PICO (Population, Intervention, Comparison, Outcome) scheme.

Other studies have exploited the entire article, for the extraction of papers' metadata as \cite{Lin:2010}: the authors propose a preliminary system based on CRF for extracting formulaic text (authors names, email and institution) as well as some key study parameters as free text, from PubMedCentral articles. They reach promising results for the formulaic text, but only moderate success for the free text attributes.
The study in \cite{abs-1708-06075} involves finding key-phrases from scientific articles and then classifying them. However, these categories are much broader (coarse-grained), e.g. `process', `task' etc., than the fine-grained categories in our task (min/max age).

A few studies have tackled the min/max age extraction problem.
Most research work on extracting information from clinical trial literature
considers ``eligibility criteria'' as a target element, which often contains min/max age 
information \cite{pmid18999067,Kiritchenko2010}. 

%However, min/max age information contained in the eligibility criteria expresses the plan for the study and may be different from the actual min/max age values of the study samples.

However, min/max age information contained in the eligibility criteria refers to the planned min/max age and may be different from the actual min/max age values of the study samples (for example: the researchers could decide to test a population of women between 20-30 years, but realistically they could gather participants only between 22 and 28 years old).
\cite{summerscales} carefully designed a number of heuristic rules to extract min/max age values of the study population from the abstracts.
We differ from this latter work as we (a) extract such information from the full articles and (b) use a machine learning approach.
In addition, we integrate the rules designed by \cite{summerscales} into our passage retrieval based IE system as a baseline.

Generally, in contrast to previous work, in this paper we a) concentrate on the extraction of population characteristics, b) use the entire article for detecting the min/max age and c) compare an unsupervised approach with a QA-based approach.

%

%%% FB: I MOVED THIS IN METHODS. I THINK IT WAS A BIT OUT OF CONTEXT IN RELATED WORKS

%\subsection{Question Answering}
%We use a Question Answering approach.
%Most recently, \emph{reading comprehension} or \emph{question answering based on context} has gained popularity within the NLP community, in particular since \cite{Rajpurkar16} 
%released a large-scale dataset (SQuAD) consisting of 100,000+ questions on a set of Wikipedia articles. We explore the QA framework to extract min/max age values--that is,
%%cast our problem as a machine comprehension task--that is, 
%the system first reads an article, then answers the questions ``what is the min/max age of the participants?''. 
%Various neural network models have been proposed for this task but these models trained on SQuAD do not work well in our scenario. We create training data
%for max/min age value extraction by leveraging the massive structured clinical study records on \emph{ClinicalTrials.gov} and train our question-answering module 
%on top of the Bi-Directional Attention Flow Network proposed by \cite{seo17}.

\subsection{Question Answering}
Most recently, \emph{reading comprehension} or \emph{question answering based on context} has gained popularity within the NLP community, in particular since \cite{Rajpurkar16} 
released a large-scale dataset (SQuAD) consisting of 100,000+ questions on a set of Wikipedia articles. 
In the medical domain, \cite{simon18} created a dataset of clinical case reports for machine
reading comprehension (CliCR). The dataset contains around 100,000 gap-filling queries on 12,000 case reports. These queries 
are created by blanking out medical entities in the \emph{learning points} sections using some heuristics.

We explore the QA framework for min/max age value extraction.
Various neural network models have been proposed for question answering but these models trained on SQuAD or CliCR do not work well in our scenario 
because these datasets do not contain the queries targeting the specific min/max age values expressed in the text. 
Therefore, we leverage instead the massive structured clinical study records on \emph{ClinicalTrials.gov} and create the training data for our min/max age value extraction
component.
%\francesca{@YOUFANG: do you have a version of the related works with those works on SQuAD? Why do not work well in our scenario?}

\subsection{Non-factual Information Detection} 
There has been a significant amount of research in detecting speculative language in scientific research articles \cite{hyland98,light04,kilicoglu08,medlock07,farkas10,morante12}.
%\cite{light04,hyland98,farkas10}. 
Our task requires to extract information from definite statements, therefore we use a list of speculation cues to filter out 
sentences where min/max age information are expressed speculatively.

\section{Approach}
\label{sec:method}
\begin{figure}[t!]
\begin{center}
\includegraphics[width=0.4\textwidth]{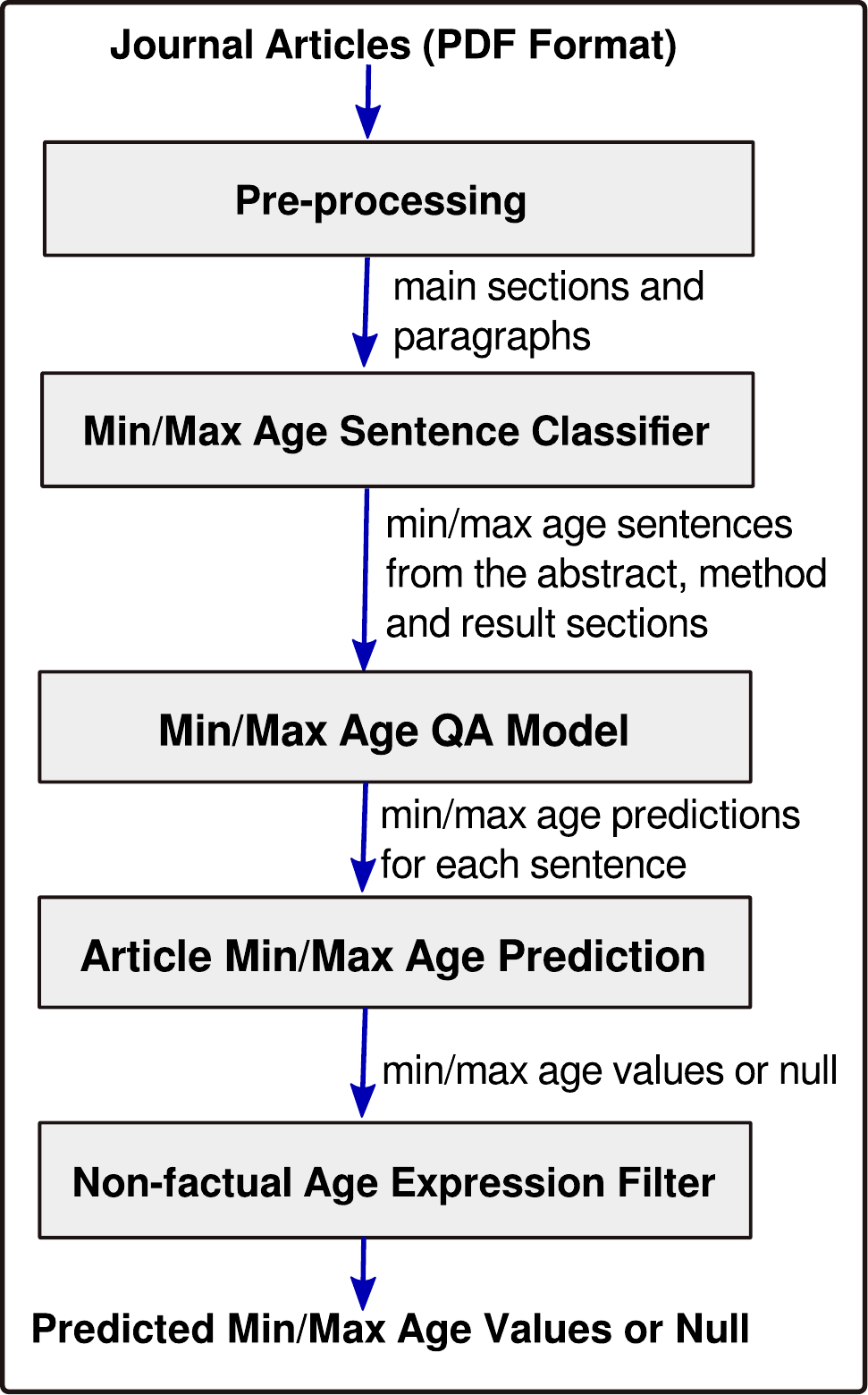}
\end{center}
\caption{Proposed QA based factual min/max age value extraction framework.}
\label{fig:system}
\end{figure}

We develop a pipeline to extract factual min/max age information from clinical trial studies. 
We divide the task in two steps: 1) finding sentences containing min/max age information; 2) extracting the value from those sentences. For the first we develop \textbf{Min/Max age Sentence Classifier} and for the second we propose a QA approach and develop the module \textbf{Min/Max Age QA Model}.

Figure \ref{fig:system} illustrated the process associated with our proposed  system. In the following sections, we describe how we create training data from \emph{ClinicalTrials.gov} as well as each component of our system in detail.

\subsection{Creating Training Data Using Clinical Study Records}

\begin{figure*}[t]
\begin{center}
\includegraphics[width=1.0\textwidth]{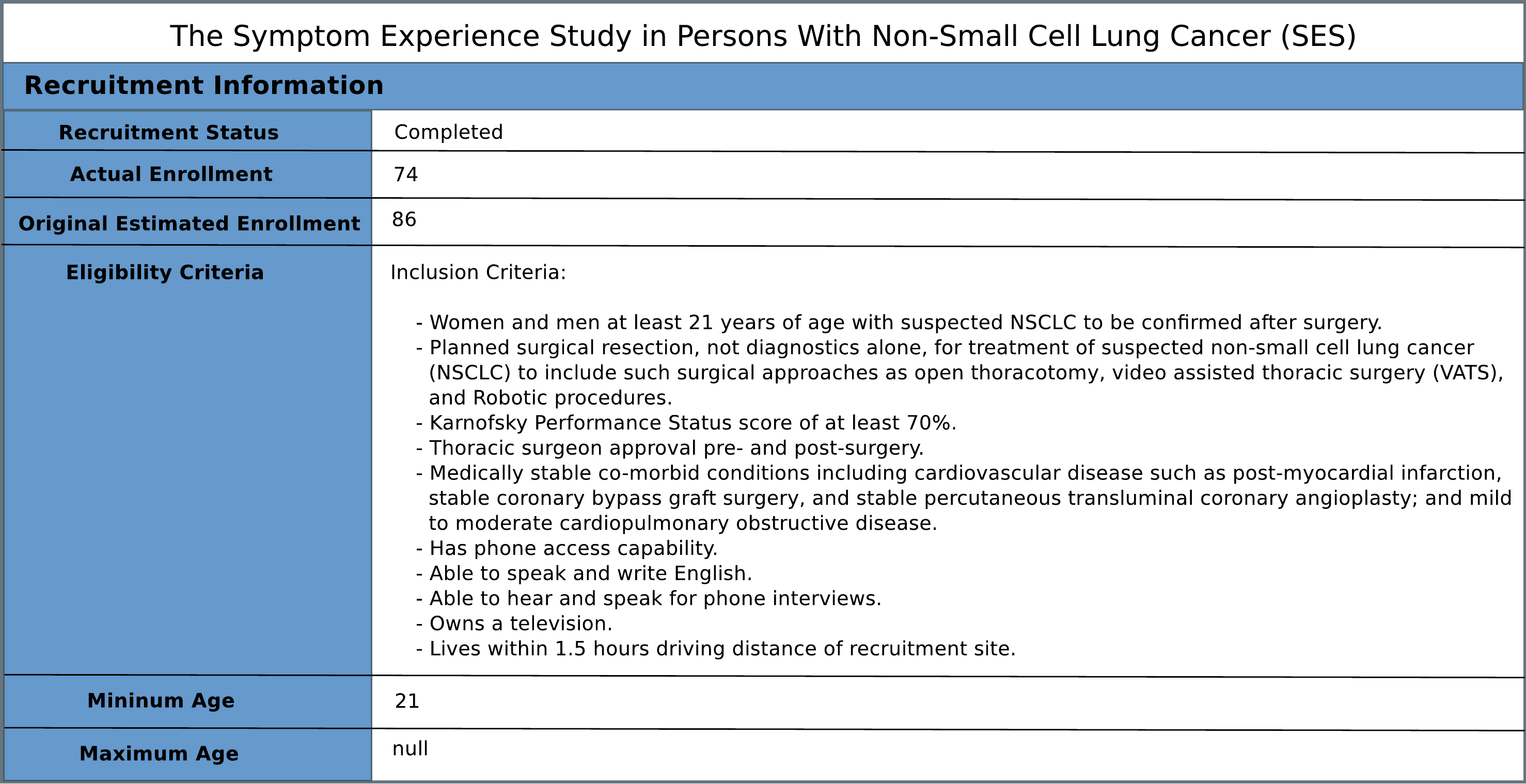}
\end{center}
\caption{An example of a clinical study record from \emph{ClinicalTrials.gov}.}
\label{fig:example}
\end{figure*}

We leverage the massive structured clinical study records on \emph{ClinicalTrials.gov} to create training data for \emph{Min/Max Age Sentence classifier} and
\emph{Min/Max Age QA Model}. \emph{ClinicalTrials.gov} is one of the largest database of clinical studies conducted around the world. It currently holds registrations around 273,000 trials from 204 countries.
Each trial registration record contains a column called ``Eligibility Criteria'', additionally min and/or max age values are indicated if they are present in the description text of the eligibility 
criteria. Figure \ref{fig:example} shows an example of a clinical study record from \emph{ClinicalTrials.gov}.

Note that most min/age expressions in eligibility criteria are speculative (e.g., \emph{at least 21 years of age}, or \emph{child must be ages 6-12 years old}), 
nevertheless they are still reflective of various linguistic forms for factual min/max age (e.g., \emph{aged 6-12 years old} or \emph{age $\geq 18$ years }). 
Therefore we expect that the models trained on this ``noisy'' dataset can still (1) identify sentences containing min/max age information and (2) predict the min/max age values.   

\subsection{Pre-processing}
Given a research article in PDF format, we first extract clean text from the PDF file using GROBID \cite{gorbid}. We associate each paragraph 
to one of the five main sections: \emph{abstract}, \emph{introduction}, \emph{method}, \emph{result} and \emph{discussion}. This step may introduce some noise (e.g., 
including the content from the table as the main body text)
because parsing PDF file in different styles %itself %is still
%remains a challenging task.
is a challenging task in itself.

\subsection{Identifying Sentences Containing Min/Max Age Information:Min/Max age Sentence Classifier}
\label{sec:agesentfinder}
After pre-processing, we identify sentences that contain min/max age information. At the inference stage, we first split paragraphs to sentences using Stanford CoreNLP Toolkit \cite{mann14}, then 
apply a classifier (\emph{MinMaxAgeSentFinder}) to predict sentences containing min/max age information among all sentences containing the word ``age/ages/aged'' or ``year/years''. 

To train our \emph{MinMaxAgeSentFinder} classifier, we create the training data using the eligibility criteria of the structured clinical study records from \emph{ClinicalTrials.gov}.
Text in eligibility criteria can be quite long (for instance, some criteria contain more than 10 clauses/sentences), so we only keep the clause/sentence which contains the annotated min/max age value(s).
More specifically, we first split the eligibility criteria into sentences/clauses using the delimiter ``-'', then choose the clauses/sentences which contain
the annotated min/max age values as well as the word ``age/ages/aged'' or ``year/years''.
For instance, in the example shown in Figure \ref{fig:example}, we will keep the sentence ``\emph{Women and men at least 21 years of age with suspected NSCLC to be confirmed after surgery.}''
as the positive training instance and filter out other sentences/clauses.

We randomly choose 20,000 such sentences/clauses (10,000 for min age and 10,000 for max age) as positive training instances. Negative training instances are sentences which do not contain the word 
``age/ages/aged'' or ``year/years'' from 60 clinical research articles. Note that these articles are different from the articles in the testing dataset. 
We use MaxEnt classifier to train \emph{MinMaxAgeSentFinder} with the following features: adjacent word n-grams (n=1-4) and adjacent 
letter n-grams within words.

\subsection{Predicting Min/Max Age Values for Each Sentence: Min/Max Age QA Model}
\label{sec:minmaxageqa}
We approach the problem of extracting values of min/max age from a question-answering perspective. Specifically, our system first reads a sentence, then answers the questions ``what is the min/max age of the participants?''.

%Most recently, \emph{reading comprehension} or \emph{question answering based on context} has gained popularity within the NLP community, in particular since \cite{Rajpurkar16} 
%released a large-scale dataset (SQuAD) consisting of 100,000+ questions on a set of Wikipedia articles. 

Various neural network models have been proposed for this task but these models trained on SQuAD do not work well in our scenario, because SQuAD does not contain this type of question-answer pairs.
Therefore we create training data
for max/min age value extraction by leveraging the massive structured clinical study records from \emph{ClinicalTrials.gov}
The training data are 10,000 $<$eligibility criteria--min age$>$ pairs and 10,000 $<$eligibility criteria--max age$>$ pairs described previously. Note that we use the whole eligibility criteria instead of choosing the specific sentence/clause which contain the min/max age value.
We believe that with the additional min/max age information, the question-answering module can locate the position of the min/max age value and learn 
various patterns for the target question.

%and train our question-answering module 
%on top of the Bi-Directional Attention Flow Network proposed by \cite{seo17}.

We train our min/max age question-answering module (\emph{MinMaxAgeQA}) using the Bi-Directional Attention Flow (BiDAF) Network \cite{seo17}.
BiDAF uses attention mechanisms in both directions (i.e., question-to-context and context-to-question) to find a sub-phrase from the input text to answer the question. 

BiDAF includes both character-level and word-level embeddings. Most word tokenization models are not robust for numeric expressions in scientific literature. For instance, 
the Stanford CoreNLP tokenizer tokenizes the clause: ``\emph{aged 6-12 years old}'' as ''\emph{\{aged, 6-12, years, old}\}'' - it does not recognize 6 and 12 as two different tokens.
The character-level embeddings in BiDAF can overcome this problem and the module correctly predicts 6 is the value of min age for this example.

\subsection{Predicting Min/Max Age Values for Each Article}
To predict min/max age values of the study samples for each article, we apply \emph{MinMaxAgeQA} to each predicted sentence containing min/max age information from 
%\emph{MinMaxAgeSentFinder} 
the \emph{abstract}, \emph{method}, and \emph{result} sections %\footnote{We decided not to include sentences from the \emph{introduction} section because it may include other min/max age information
%which is not related to the study samples (see Example \ref{ex:age5} and Example \ref{ex:age6}). We leave filtering out unrelated min/max age information from introductions as future work.}
%Future work will use other method to filter out unrelated min/max age information from introductions.}
%We do not include sentences from the \emph{introduction} section because it may talk about other min/max age information which is not related to the study samples (see Example \ref{ex:age5} and Example \ref{ex:age6}).
%}
on both questions (i.e., \emph{what is the min age of the participants?} and \emph{what is the max age of the participants?}). 
Answers that do not represent a valid integer number or answers whose confidence score are less than 0.5 are discarded. 
For each question, we keep the answer with the highest confidence score.

We do not include sentences from the \emph{introduction} section because it may include other min/max age information
which is not related to the study samples (see Example \ref{ex:age5} and Example \ref{ex:age6}). We leave filtering out unrelated min/max age information from introductions as future work.

\begin{figure}[t]
\begin{center}
\includegraphics[width=0.45\textwidth]{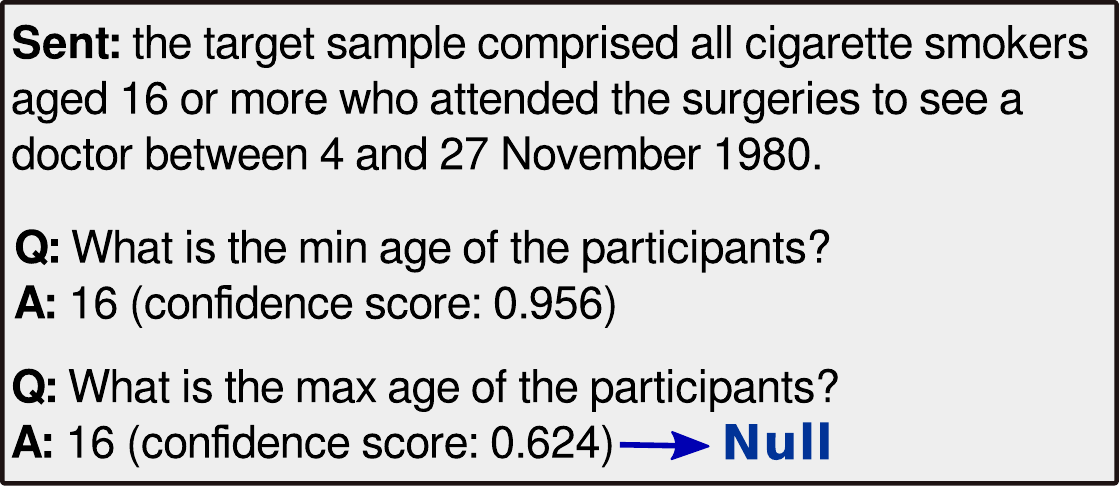}
\end{center}
\caption{Conflicting min/max age values.}
\label{fig:ageinference}
\end{figure}

Finally, if both min and max age values are predicted for an article, we check whether the min age value is smaller than the max age value. Otherwise we keep 
the answer with the higher confidence score and discard the other one. For instance, as shown in Figure \ref{fig:ageinference}, the number 16 is predicted as both 
the min age value (with the probability of 0.956) and the max age value (with the probability of 0.624) for an article, we keep 16 as the prediction for the min age value 
and set the prediction of the max age to "Null``.

%We decide not to include sentences from the introduction section because, in our corpus, it often includes other min/max age information which is not related to the study samples ( see Example 5 and Example 6) . We are aware that this involve a loss of information. Future work will address other methodology to filter out unrelated min/max information from introductions.

\subsection{Non-factual Age Expression Filter}
\label{sec:filter}
In this component, we filter out a min/max age value prediction if it is expressed speculatively. 
We first extract the clause which contains the prediction, then check whether a speculation cue word/phrase is present in the clause using the speculation cues from \cite{light04}.
These cue words are: \emph{\{if, at least, must, had to, has to, have to, need, needs\}}.

\section{Evaluation}\label{sec:eval}
\subsection{Testing Dataset}

\begin{table*}[t]
\begin{center}
\begin{tabular}{l|l}
\hline
 \multicolumn{2}{c}{\emph{Testing Dataset}} \\ \hline
 \emph{\# of articles} & 50\\ \hline
\emph{\# of sentences} & 18,417\\ \hline
\emph{\# of tokens (main text)} & 432,056\\ \hline
\emph{\# of sentences containing } & 843\\ 
\emph{``age/ages/aged'' or ``year/years''} & \\ \hline
\emph{\# of numeric tokens in sentences containing } & 2,226\\ 
\emph{``age/ages/aged'' or ``year/years''} & \\ \hline
\end{tabular}
\end{center}
\caption{Statistics for the testing dataset.}
\label{tab:testset}
\end{table*}

The ground-truth dataset used for evaluation comprises a set of 50 published journal articles in PDF format on smoking cessation. 
The dataset contains around 432k tokens and 18k sentences. Table \ref{tab:testset} shows some statistics about the testing dataset.
Overall, we have 843 sentences containing the word ``age/ages/aged'' or ``year/years'' and these sentences contain 2,226 numeric tokens.

The articles were
annotated by a team of four behaviour science domain experts in the context of a broader project focused on leveraging the scientific literature in behaviour 
change \cite{michie2017human}.
Annotation for a
particular document was performed by two human annotators using the EPPI tool\footnote{\url{http://eppi.ioe.ac.uk/CMS/}}. The annotation process involved highlighting relevant pieces of text and then assigning them to
the corresponding min/max age attribute.  Additionally, in order to disambiguate the highlighted text, the annotators were asked
to annotate the entire sentence containing the highlighted piece as the additional context. Conflicts in the annotation process
were resolved through discussions. Note that not every document contains a min/max age annotation.
This is because not every article reports the min/max age of the overall study samples. 
In the testing dataset, 35 papers have min age annotations and 25 papers have max age annotations.

\subsection{Evaluation Metric}
We use recall, precision and F-score for evaluation. Recall is calculated as the number of articles
where the min/max age values are correctly predicted divided by the number of articles where min/max age values
are annotated. Precision is calculated as the number of articles where the min/max age values are correctly predicted 
divided by the number of articles where the system makes a min/max age value prediction. F-score is the harmonic average of the precision and recall.

\subsection{Baseline 1: \emph{PassageRetrievalBasedMinMaxAgeExtractor}}
We developed a passage retrieval based IE system to extract min/max age 
values \cite{ganguly2018unsupervised}.
The first step is to retrieve the passages containing 10, 20, and 30 words using
the query ``\emph{(age OR ages OR aged OR year OR years)}''.  The intention of
retrieving passages is to restrict extraction of factoid answers to potentially relevant small semantic units of text rather
than the text of the whole document. 

The next step is to use validation criteria to select the likely answer candidates. We use the min/max age 
patterns
from \cite{summerscales} as the validation criteria to choose the likely answer candidates from each retrieved passage for min age and max age respectively.
These patterns can be viewed as rules which are carefully designed by humans to extract min/max age values.
For instance, a rule can be: \emph{if a passage contains the phrase ``\textbf{greater than X}'' or ``\textbf{older than X}'' and \emph{X} is an integer number between 10 to 100, 
then choose 
\emph{X} as an answer candidate}. 
It is worth noting that \cite{summerscales} is the only previous work targeting the same task according to our best knowledge. 
We integrate all the heuristic rules for min/max age value extraction from \cite{summerscales} into our passage retrieval based IE system.

Finally, we score the answer candidates by a term proximity function that
takes into account the differences in position between the query terms and the candidate answers \cite{Zhao2009}.
The function is formally defined in the following Equation:
\begin{equation}
sim(c,Q) = \frac{1}{\vert Q \vert}\sum_{q \in Q}exp(-(p_c-p_q)^2/\sigma)  
%A=\{a: \phi(p,q,a,\tau)=1\}
\label{eq:plm}
\end{equation}
Equation \ref{eq:plm} describes the proximity based ranking function between a candidate answer $c$ and a query $Q$, denoted by $sim(c,Q)$.
Practically, for each word in the passage that matches the query terms ($q$), the similarity function increases the score of that candidate 
by an amount that depends on the distance between that matched word and the candidate answer $(p_c-p_q)$.
Specifically, we use a Gaussian function centered at each query term  to determine the 
increase in similarity score. The parameter $\sigma$ controls the bandwidth of the Gaussians and is set to $1$ in our experiments.

\subsection{Baseline 2: \emph{CRFBasedMinMaxAgeExtractor}}
We also developed the second baseline using CRF \cite{sutton12}.
The training dataset contains the clauses/sentences which contains the annotated min/max age value(s) from the eligibility criteria of the 
clinical studies registered in \emph{ClinicalTrials.gov}. For each clause/sentence, we use Stanford CoreNLP Toolkit \cite{mann14} to obtain the tokens 
as well as the POS tags, then we create the corresponding training instance using BIO labels (i.e., Beginning/Inside/Outside of a min/max age).
Table \ref{tab:crfexample} shows the training instance for the example illustrated in Figure \ref{fig:example}.  %for the min age CRF model.

\begin{table}[h]
\begin{center}
\begin{tabular}{l|l|c}
\hline
 Token & POS tag & MinAgeAnnotation\\ \hline
 Women & NNS & O\\ 
 and & CC & O\\ 
 men & NNS& O\\ 
 at & IN & O\\ 
 least& JJS& O\\ 
 21 & CD & B\\ 
 years & NNS& O\\ 
 of & IN & O\\ 
 age & NN& O\\ 
 with & IN& O\\ 
 suspected & VBN& O\\ 
 NSCLC & NNP& O\\ 
 \ldots & \ldots & \ldots\\ \hline 
\end{tabular}
\end{center}
\caption{A training instance for the min age extraction CRF model.}
\label{tab:crfexample}
\end{table}

%50 papers
\begin{table*}[t]
\begin{center}
\begin{tabular}{|l|l|l|l|l|l|l|}
\hline
 &  \multicolumn{3}{c|}{\emph{MinAge}} &\multicolumn{3}{c|}{\emph{MaxAge}} \\ 
 & R & P & F & R & P & F\\ \hline
{\emph{Baseline 1: PassageRetrievalBasedMinMaxAgeExtractor}} & 28.6&90.9&43.5&32.0&72.7&44.4\\ \hline
{\emph{Baseline 2: CRFBasedMinMaxAgeExtractor}} & 50.0&42.5&45.9&25.0&18.2&21.1\\ \hline
{\textbf{This work: QABasedMinMaxAgeExtractor}} & 65.7&79.3&\textbf{71.9}&60.0&75.0&\textbf{66.7}\\

\hline
\end{tabular}
\end{center}
\caption{Experimental results. Bold
indicates statistically significant differences over the
baseline using randomization test (p $<$ 0.01).}
\label{tab:results}
\end{table*}

\begin{table*}[t]
\begin{center}
\begin{tabular}{|l|l|l|l|l|l|l|}
\hline
 &  \multicolumn{3}{c|}{\emph{MinAge}} &\multicolumn{3}{c|}{\emph{MaxAge}} \\ 
 & R & P & F & R & P & F\\ \hline

{\emph{This work: QABasedMinMaxAgeExtractor}} & 65.7&79.3&\textbf{71.9}&60.0&75.0&\textbf{66.7}\\
{\emph{---WO MinMaxAgeSentFinder}} & 68.6&68.6&\textbf{68.6}&52.0&56.5&54.2\\
{\emph{---WO MinMaxAgeQA}} &31.4 &84.6&45.8&40.0&71.4&51.3\\
{\emph{---WO Non-factualSentFilter}} & 68.6&70.6&\textbf{69.6}&60.0&71.4&\textbf{65.2}\\
\hline
\end{tabular}
\end{center}
\caption{Contribution of each component to the overall system performance.}
\label{tab:results-extra}
\end{table*}

We train two CRF models for min age and max age extraction respectively, using 10,000 training instances for each model.
We use words as well as POS tags as features. More specifically, for the word type features, we consider the current word $w_i$, the surrounding words 
($w_{i-2}$, $w_{i-1}$, $w_{i+1}$, $w_{i+2}$), as well as bi-grams ($w_{i-1}$ + $w_{i}$, $w_{i}$ + $w_{i+1}$) and tri-grams 
($w_{i-2}$ + $w_{i-1}$ + $w_{i}$, $w_{i-1}$ + $w_{i}$ + $w_{i+1}$, $w_{i}$ + $w_{i+1}$ + $w_{i+2}$) created from words. We create similar unigram, bi-gram and tri-gram features
using the automatically predicted POS tags as well. We also include the combinations of the previous prediction and the current prediction as bi-gram features.

At the inference stage, for each article, we first extract all sentences containing the word ``age/ages/aged'' or ``year/years''.
We then apply the min/max age CRF model on these sentences and extract all tokens with the predicted label ``B''.
In the end, among all predicted words, we choose the word which represents a valid integer number and has the highest confidence score as 
the predicted min/max age value for the article.

\subsection{Results and Discussion}
Table \ref{tab:results} shows the performance of the baselines (\emph{PassageRetrievalBasedMinMaxAgeExtractor} and \emph{CRFBasedMinMaxAgeExtractor}) 
as well as our system (\emph{QABasedMinMaxAgeExtractor}, described in Section \ref{sec:method}) for extracting min/max age values of the study samples. 

For \emph{MinAge}, the first baseline (\emph{PassageRetrievalBasedMinMaxAgeExtractor}) achieves a very high precision score ($90.9\%$) but suffers from low recall ($28.6\%$). 
The second baseline (\emph{CRFBasedMinMaxAgeExtractor}) improves the recall by 21.4 points but only achieves a precision score of 42.5\%.
Compared to the first baseline, our system manages to improve recall by 37.1 points %\% 
%from 28.6\% to 65.7\%  
and still achieves a reasonable level of precision ($79.3\%$). Overall, 
our system improves the results over the two baselines by a large margin %in terms of 
regarding F-score ($71.9\%$ vs. $43.5\%$, and $71.9\%$ vs. $45.9\%$). 

The similar pattern is also observed for \emph{MaxAge}: Our system improves the results over the first baseline by a substantial margin on 
recall ($60.0\%$ vs. $32.0\%$) and F-score ($66.7\%$ vs. $44.4\%$) respectively.

It might seem surprising that \emph{CRFBasedMinMaxAgeExtractor} performs much worse than \emph{PassageRetrievalBasedMinMaxAgeExtractor} for \emph{MaxAge}.
This is because  many max age values in scientific articles are not correctly recognized as a single token by Stanford CoreNLP Toolkit.
For instance, the tokenization model predicts that ``\emph{18-60}'' or ``\emph{$<=$60}'' as single tokens. 
In contrast, our system is more robust for parsing such numeric expressions. 

In addition, it seems that the carefully designed min/max age patterns in the first baseline only cover a few forms of min/max age expressions.
On the contrary, our min/max age question-answering module (\emph{MinMaxAgeQA}, Section \ref{sec:minmaxageqa}) trained over
a large-scale dataset can capture various linguistic expressions of min/max age in natural language, for instance, ``\emph{$\geq 18$ years of age}''
 or ``\emph{age $>=$ 18 years}''.

%\subsection{Ablation Experiment} 
\subsection{Analysis} 
To better understand the roles of different components in our system, we carried out a few experiments:
\begin{itemize}
\item \emph{---WO MinMaxAgeSentFinder}: instead of using \emph{MinMaxAgeSentFinder} to find the sentences containing min/max age information, we pass all sentences 
 containing the word ``age/ages/aged'' or ``year/years'' from the abstract, method, and result sections to the next component \emph{MinMaxAgeQA}.
\item \emph{---WO MinMaxAgeQA}: we use the most common min/max age expression pattern in clinical trial studies ``\emph{X-Y}'' (e.g., \emph{18-23 years old}) to predict min and max age values from the 
first sentence contain such a pattern.
\item \emph{---WO Non-factualSentFilter}: Non-factual age expression filter is not used.
\end{itemize}

The results of these experiments are shown in Table \ref{tab:results-extra}. It seems that \emph{MinMaxAgeQA} has the most impact on the performance while \emph{Non-factualSentFilter} has less of an impact. 
In addition, \emph{MinMaxAgeSentFinder} has more impact on the results of \emph{MaxAge} compared to \emph{MinAge}.

We also performed some error analysis on our full system. We noticed that the noise introduced in the pre-processing step (e.g., missing some paragraphs)
is the main reason to cause our system to predict ``Null'' for articles with min/max age annotation. For cases where a wrong min/max age value is predicted, 
they are often embedded in the speculative expressions which are not captured by our current \emph{Non-factualSentFilter}. For instance, 
the system predicts
\textbf{24} as the max age for one article in which \textbf{24} appears in a speculative sentence (see \emph{speculative expression} in Example \ref{ex:age7}).
For this article, the annotation for max age is \textbf{23} (see \emph{factual expression} in Example \ref{ex:age7}).

\begin{examples}
\item \label{ex:age7} (\emph{speculative expression}) Eligibility for this study included being a student (full or part time), smoking at least 1 cigarette/day in each of the past 7 days, being aged 18-\textbf{24} years, and being interested in quitting smoking in the next 6 months.
\\
(\emph{factual expression}) Participants were 83 smokers, who were 18-\textbf{23} years old and undergraduate students at a university.
\end{examples}

\section{Conclusions}\label{sec:con}
This paper aims to extract factual min/max age values of the study samples from clinical research papers.
We leverage the large-scale records from the \emph{ClinicalTrials.gov} database to provide distant supervision for our system. 
We also explore ``speculative cues'' and the structure of the scientific papers to extract information from factual statements about the target study.
We show that our approach outperforms a passage retrieval based IE system and a CRF-based model by a large margin on a testing dataset
consisting of 50 journal articles and around 18,000 sentences.

In the future, we plan to extend our framework to extract other types of numeric values from the clinical research papers, such as the outcome values 
of the different intervention groups and the control group (e.g., \emph{40\% of PP abstinence rates}), as well as the time frame of the follow up (e.g., \emph{52 weeks} or \emph{6 months}).

\section*{Acknowledgments}
This  work  was  supported  by  a  Wellcome  Trust  collaborative  award  as  a  part  of  the  Human Behaviour-Change Project (HBCP): Building the science of behaviour change for complex intervention development (grant no. 201,524/Z/16/Z).

\bibliographystyle{acl_natbib}
\bibliography{bib/lit}

\end{document}